\documentclass[11pt]{article}

\usepackage[preprint]{acl}

\usepackage{times}
\usepackage{latexsym}

\usepackage[T1]{fontenc}
\usepackage[utf8]{inputenc}

\usepackage{microtype}

\usepackage{inconsolata}

\usepackage{graphicx}
\usepackage{amsmath}

\usepackage{hyperref}
\usepackage{url}
\usepackage{booktabs}
\usepackage{multirow}
\usepackage{enumitem}
\usepackage{subcaption}
\usepackage{wrapfig}
\usepackage{xcolor}

\usepackage[most]{tcolorbox}
\definecolor{darkgreen}{RGB}{0,50,0}
\tcbset{
    promptbox/.style={
        enhanced,
        colback=gray!10,
        colframe=darkgreen!70,
        colbacktitle=darkgreen!75,
        coltitle=white,
        boxrule=0.8pt,
        arc=2.5mm,
        left=6pt,
        right=6pt,
        top=6pt,
        bottom=6pt,
        fonttitle=\bfseries,
    }
}

\title{The Anatomy of Uncertainty in LLMs}

\author{
Aditya Taparia$^{1}$ \thanks{Correspondence to: Aditya Taparia <ataparia@asu.edu>.} \quad
Ransalu Senanayake$^{1}$ \quad
Kowshik Thopalli$^{2}$ \quad
Vivek Narayanaswamy$^{2}$ \\
\\
$^{1}$School of Computing and Augmented Intelligence, Arizona State University \\
$^{2}$Lawrence Livermore National Laboratory
}

\begin{document}
\maketitle
\begin{abstract}
Understanding \textit{why} a large language model (LLM) is uncertain about the response is important for their reliable deployment. Current approaches, which either provide a single uncertainty score or rely on the classical aleatoric-epistemic dichotomy, fail to offer actionable insights for improving the generative model. Recent studies have also shown that such methods are not enough for understanding uncertainty in LLMs. In this work, we advocate for an uncertainty \textit{decomposition} framework that dissects LLM uncertainty into three distinct \emph{semantic} components: (i) input ambiguity, arising from ambiguous prompts; (ii) knowledge gaps, caused by insufficient parametric  evidence; and (iii) decoding randomness, stemming from stochastic sampling. Through a series of experiments we demonstrate that the dominance of these components can shift across model size and task. 
Our framework provides a better understanding to audit LLM reliability and detect hallucinations, paving the way for targeted interventions and more trustworthy systems. 
\textbf{Code:} \href{https://github.com/aditya-taparia/LLM-Uncertainty}{https://github.com/aditya-taparia/LLM-Uncertainty}
\end{abstract}

\section{Introduction}

Large Language Models (LLMs) have achieved remarkable success in complex reasoning and generation tasks. Despite their great capabilities, they have the tendency to generate plausible-sounding but uncertain responses. Understanding when and why these models are uncertain in their response helps in detecting hallucination~\cite{manakul2023selfcheckgpt, kadavath2022language, kuhn2023semantic}, improving response quality~\cite{ramirez2024optimising}, and optimizing tool calling~\cite{zubkova2025sugar}. 
% Despite rapid advancements, this phenomenon persists as a critical barrier for their reliable deployment.  
Recent studies have shown that hallucinations in LLMs are triggered because the model often guesses when they are unsure about the final response~\cite{kalai2025language}. This makes it important to identify when the model is uncertain and the fundamental nature and origins of uncertainty. 

Uncertainty in LLMs can originate from different sources. Consider the case of Gemma3 27B model~\cite{team2025gemma}, when asked a straightforward question from TriviaQA~\cite{joshi2017triviaqa},
\begin{quote}
    \textit{``What was Walter Matthau's first movie?''}
\end{quote}
the model consistently responded with \textit{``The Gangster,''} while the correct answer is \textit{``The Kentuckian.''} This discrepancy highlights two possible scenarios. First, the phrasing of the question introduces input ambiguity, since “first movie” could mean Matthau’s first credited role or his earliest on-screen appearance. Second, the model’s internal knowledge may be incomplete or imprecise, reflecting knowledge gaps in its training data. Similarly, another source of discrepancy in the response could be introduced during output decoding. If we look at another example from the same dataset,
\begin{quote}
    \textit{``In Hanna and Barbera's TV cartoons base on The Addams Family who was the voice of Gomez?''}
\end{quote}
the Gemma3 27B model consistently gives correct answer, \textit{``John Astin''} when responses are generated using greedy decoding. But when temperature decoding is used, it sometimes responds with incorrect answer, \textit{``Ted Cassidy.''} 
% This shows that decoding randomness can further affect the generation from the model. 
Recent work has also demonstrated that decoding strategies influence both model outputs and the resulting uncertainty estimates~\cite{hashimoto2025decoding}. 

\begin{figure*}[t]
\centering
\includegraphics[width=0.95\linewidth]{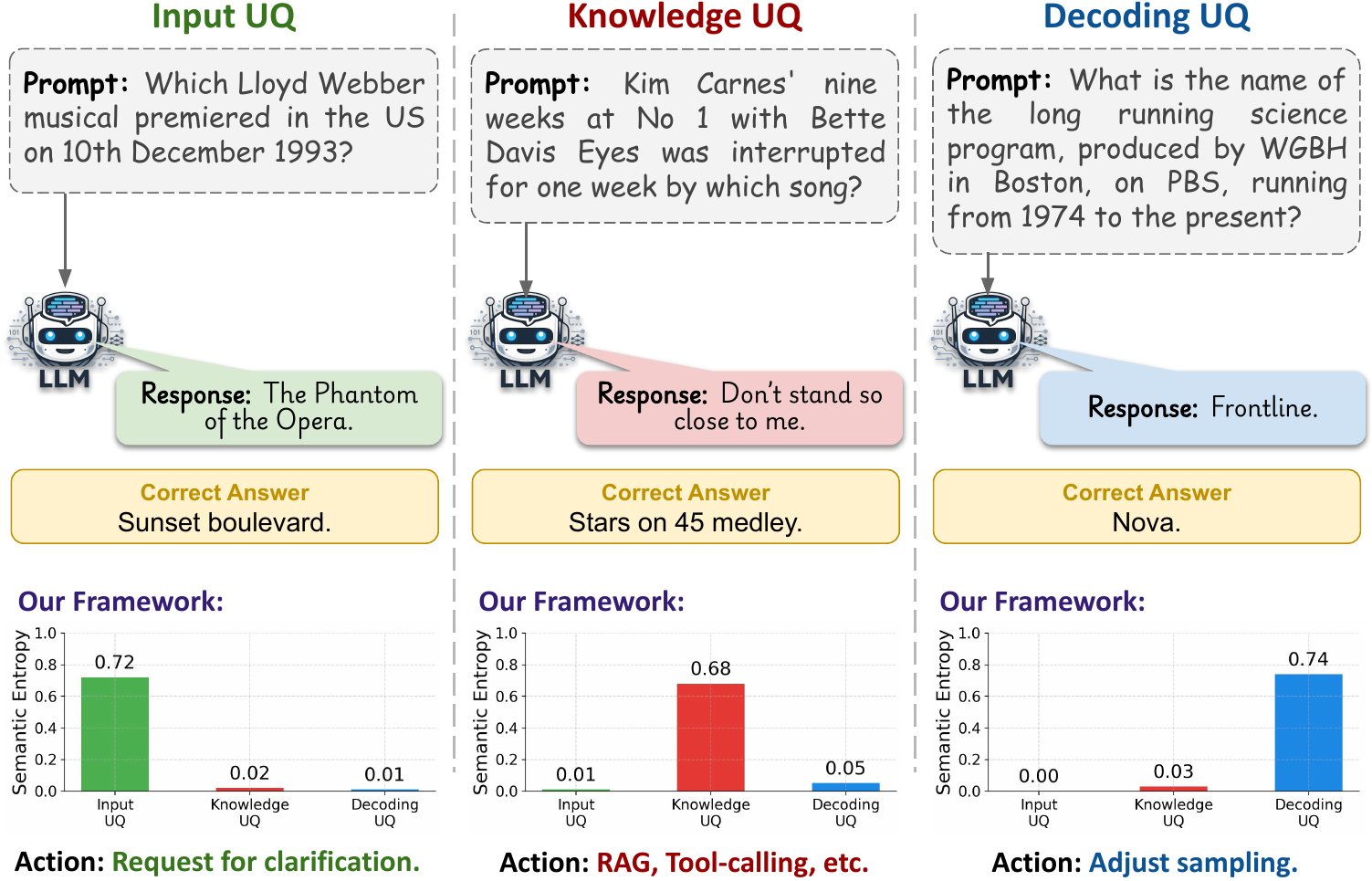}
\caption{Uncertainty decomposition on TriviaQA examples of Gemma 3 27B model. Our framework identifies the dominant source of uncertainty---input ambiguity (left), knowledge gaps (middle), or decoding randomness (right)---and maps each to a targeted mitigation action. 
}
\label{fig:intro_fig}
% \vspace{-5pt}
\end{figure*}

Existing approaches focuses on quantifying these uncertainty using a single score~\cite{manakul2023selfcheckgpt, kadavath2022language, kuhn2023semantic}. While these scores are useful for ranking responses and guiding abstention, they are not actionable because they fail to diagnose the root cause of uncertainity. It remains unclear what intervention might reduce the uncertainty in these systems. Some works have shown how these uncertainties can be decomposed into classical aleatoric-epistemic dichotomy~\cite{Senanayake2024arxiv_unc, hou2024decomposing}. However, recent studies~\cite{kirchhof2025position, huang2024survey, xie2025empirical, bakman2025reconsidering} have highlighted that this dichotomy is not sufficient in case of LLMs. 

To address these issues, in this work we propose a framework that decomposes uncertainty in LLMs into three distinct \emph{semantic} components: (i) input ambiguity, stemming from prompts with multiple valid interpretations; (ii) knowledge gaps, caused by insufficient training coverage or outdated information; and (iii) decoding randomness, introduced by the sampling process itself (see Figure~\ref{fig:intro_fig}). We advocate that this decomposition provides a more faithful description of uncertainty in generative models and offers actionable insights for system design. For example, high input-driven uncertainty suggests clarifying questions; high knowledge uncertainty suggests retrieval or data augmentation; and high decoding uncertainty suggests adjusting sampling strategy. Our contributions are:

\begin{enumerate}
    \item We propose a framework for decomposing LLM uncertainty into three distinct \emph{semantic} components: input ambiguity, knowledge gaps, and decoding randomness.
    \item We empirically demonstrate how the dominant source of uncertainty shifts across different tasks and model scales.
    % \item How we can use them to actively adapt the system. 
\end{enumerate}

% By analyzing these components individually, we demonstrate that the dominant source of uncertainty shifts depending on the input (as illustrated in Fig.~\ref{??}), providing a more granular and actionable understanding of model behavior.

% These instructions are for authors submitting papers to *ACL conferences using \LaTeX. They are not self-contained. All authors must follow the general instructions for *ACL proceedings,\footnote{\url{http://acl-org.github.io/ACLPUB/formatting.html}} and this document contains additional instructions for the \LaTeX{} style files.

% The templates include the \LaTeX{} source of this document (\texttt{acl\_latex.tex}),
% the \LaTeX{} style file used to format it (\texttt{acl.sty}),
% an ACL bibliography style (\texttt{acl\_natbib.bst}),
% an example bibliography (\texttt{custom.bib}),
% and the bibliography for the ACL Anthology (\texttt{anthology.bib}).

\section{Related Works}
\paragraph{Uncertainty Quantification in LLMs.}
Prior work has developed various techniques to assign uncertainty scores to LLM outputs. \cite{manakul2023selfcheckgpt} propose SelfCheckGPT, a sampling-based method that compares multiple model generations. They show that, if a fact is truly known, the samples agree, whereas hallucinated facts cause divergent answers. Similarly, \cite{kadavath2022language} uses LLM itself to estimate the probability that their own answers are correct (a ``P(True)'' confidence). Another approach by \cite{kuhn2023semantic} defines a semantic entropy score that accounts for linguistic paraphrases (shared meaning) to better predict uncertainty. These uncertainty scores have been useful for improving tasks like hallucination detection and inference efficiency. For instance,~\cite{ramirez2024optimising} show that using a small model’s uncertainty to decide whether to invoke a larger model yields an effective two-tier cascade. But they do not explain why the LLM is uncertain about a particular response and how we can improve them. With our approach, we aim to bridge this gap by decomposing uncertainty into interpretable sources.

\paragraph{Towards Decomposing Uncertainty in LLMs.}
Decomposing uncertainty into meaningful components has a long history in Bayesian and reinforcement learning~\cite{charpentier2022disentangling}. Inspired by this, recent work has begun exploring uncertainty decomposition in large language models. ~\cite{hou2024decomposing} introduce an input-clarification ensembling framework that generate multiple disambiguated versions of each prompt and ensemble the outputs. However, recent researches note that the simple aleatoric-epistemic split is not ideal for LLMs. \cite{kirchhof2025position} argue that classical definitions of aleatoric vs. epistemic uncertainty ``contradict each other and lose their meaning'' in open-ended, interactive language tasks. In particular, assigning fixed aleatoric and epistemic scores to each output cannot capture the nuanced, multi-turn uncertainty arising from under specified prompts. Motivated by these observations, we take a more fine-grained view and explicitly separate input ambiguity, knowledge gaps, and decoding randomness as distinct uncertainty sources.

% \section{Engines}

% To produce a PDF file, pdf\LaTeX{} is strongly recommended (over original \LaTeX{} plus dvips+ps2pdf or dvipdf).
% The style file \texttt{acl.sty} can also be used with
% lua\LaTeX{} and
% Xe\LaTeX{}, which are especially suitable for text in non-Latin scripts.
% The file \texttt{acl\_lualatex.tex} in this repository provides
% an example of how to use \texttt{acl.sty} with either
% lua\LaTeX{} or
% Xe\LaTeX{}.

\section{Anatomy of Uncertainty in LLMs}
We propose a framework for decomposing the uncertainty in a LLM's response into three distinct \emph{semantic} sources: input ambiguity, knowledge gaps, and decoding randomness. These sources correspond to the main stages of the generation pipeline: the user prompt (Input), the model parameters (Knowledge), and the generation procedure (Decoding).

Let $x \in X$ denote an input, $y \in Y$ a generated response, $\theta$ the model parameters of the LLM, and $\tau$ a decoding strategy. The model induces a conditional distribution $p(y \mid x, \theta, \tau)$. The overall predictive uncertainty for an input $x$ may be viewed as the entropy of this output distribution,
\begin{equation}
U_{\text{total}}(x) = \mathcal{H}\left(p(Y \mid x, \theta, \tau)\right).
\end{equation}
Although this score can flag uncertain outputs, it is not diagnostic. To identify \emph{why} the model is uncertain, we instead isolate one source of variation at a time while holding the others fixed, and compute semantic entropy over the resulting set of responses.

\begin{figure}[t]
    \centering

    % Dummy light grey placeholder box
    % \colorbox{gray!15}{%
    %     \parbox[c][7cm][c]{0.9\linewidth}{%
    %         \centering
    %         Input UQ
    %     }%
    % }

    % Uncomment this later to use the actual image instead
    \includegraphics[width=0.95\linewidth]{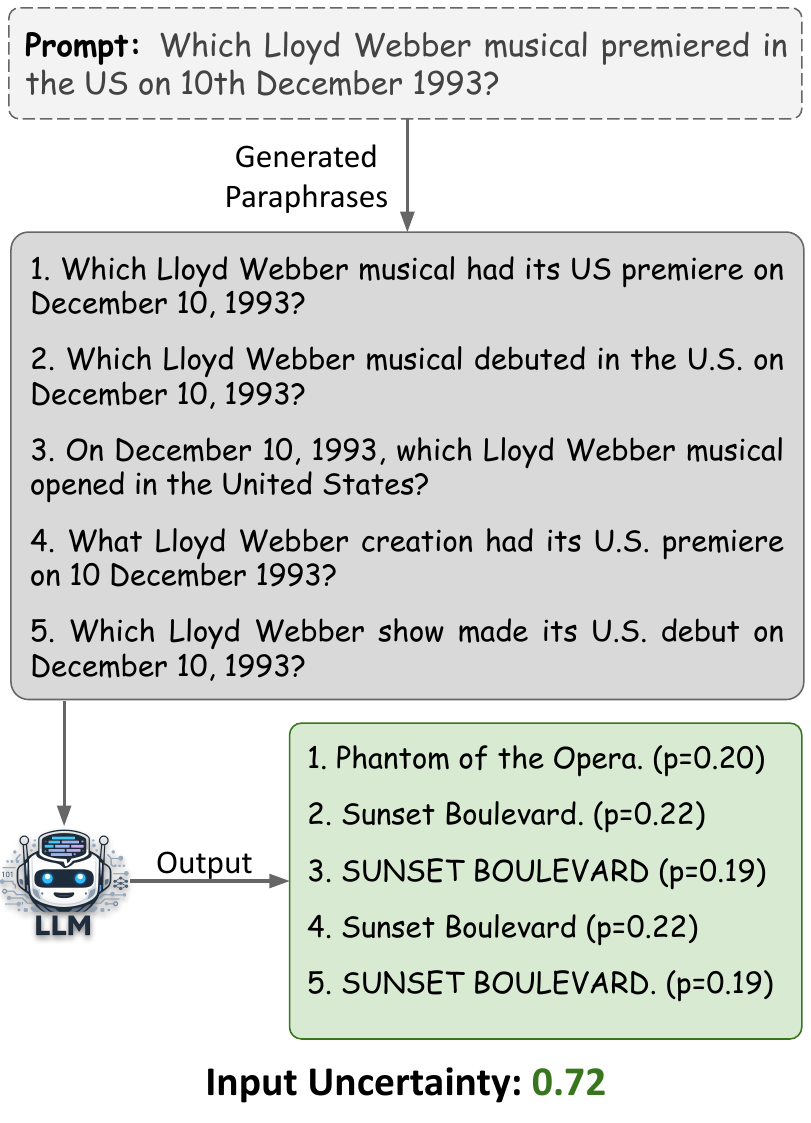}

    \caption{Illustration of input ambiguity estimation. We generate semantically equivalent paraphrases of the original question, obtain one response for each paraphrase using the same model and decoding policy, and group the responses into semantic clusters.}
    \label{fig:input_uq_example}
\end{figure}

\paragraph{Input Ambiguity (\texorpdfstring{$U_{\text{input}}$}{U-input}).}
Input ambiguity captures uncertainty caused by how the prompt is phrased. As shown in Fig.~\ref{fig:input_uq_example}, to isolate this source, we fix $\theta$ and $\tau$, and consider a set of $K$ meaning-preserving paraphrases,
\[
P(x)=\{x^1, x^2, \dots, x^K\},
\]
where each variation preserves the intent of the original query rather than introducing arbitrary textual corruption. We generate one response for each paraphrase and group the resulting outputs into semantic equivalence classes. Let $C$ denote the set of such semantic clusters. We define input-induced uncertainty as,

\begin{equation}
\begin{aligned}
U_{\text{input}}(P,\theta,\tau)
&= -\sum_{c \in C} \hat{p}(c)\log \hat{p}(c), \\
\text{where}\quad
\hat{p}(c)
&= \frac{\sum_{y^k \in c} p(y^k \mid x^k,\theta,\tau)}
{\displaystyle \sum_{c' \in C}\sum_{y^k \in c'}
p(y^k \mid x^k,\theta,\tau)}
\end{aligned}
\label{eq:input_uq}
\end{equation}

High $U_{\text{input}}$ indicates that semantically equivalent phrasings lead to different meanings in the model outputs, suggesting that the query is underspecified and may benefit from clarification.

\begin{figure}[t]
    \centering

    % Dummy light grey placeholder box
    % \colorbox{gray!15}{%
    %     \parbox[c][7cm][c]{0.9\linewidth}{%
    %         \centering
    %         Knowledge UQ
    %     }%
    % }

    % Uncomment this later to use the actual image instead
    \includegraphics[width=0.95\linewidth]{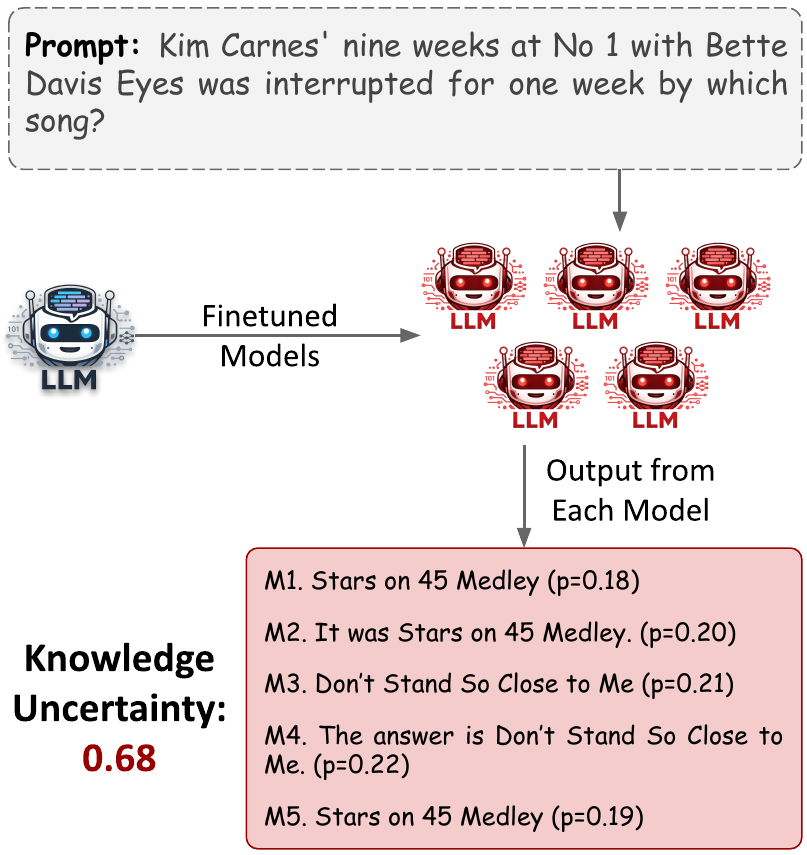}

    \caption{Illustration of knowledge-gap uncertainty estimation. For a fixed input and decoding policy, we query an ensemble of LoRA-adapted model realizations and group the resulting responses into semantic clusters. Disagreement across ensemble members reflects uncertainty arising from parametric knowledge gaps.}
    \label{fig:knowledge_uq_example}
\end{figure}

\paragraph{Knowledge Gaps (\texorpdfstring{$U_{\text{knowledge}}$}{U-knowledge}).}
Knowledge uncertainty captures variability arising from the model parameters. Since exact posterior inference over LLM parameters is intractable, we approximate multiple plausible model realizations using an ensemble,
\[
\Theta = \{\theta^1, \theta^2, \dots, \theta^M\},
\]
where each $\theta^m$ is obtained through a distinct LoRA adaptation over the training set. As shown in Fig.~\ref{fig:knowledge_uq_example}, by fixing $x$ and $\tau$, we generate one response from each realization and compute semantic entropy over the resulting outputs:
\begin{equation}
\begin{aligned}
U_{\text{knowledge}}(x, \Theta, \tau)
&= -\sum_{c \in C} \hat{p}(c)\log \hat{p}(c), \\
\text{where}\quad
\hat{p}(c)
&= \frac{\sum_{y^m \in c} p(y^m \mid x, \theta^m, \tau)}
{\displaystyle\sum_{c' \in C}\sum_{y^m \in c'} p(y^m \mid x, \theta^m, \tau)}
\end{aligned}
\label{eq:model_uq}
\end{equation}
High $U_{\text{knowledge}}$ reflects disagreement across model realizations and serves as a practical proxy for parametric knowledge gaps.

\paragraph{Decoding Randomness (\texorpdfstring{$U_{\text{dec}}$}{U-dec}).}
Decoding uncertainty captures variability introduced by the response generation procedure itself. Let $T$ denote a family of decoding strategies used in real deployment, such as greedy decoding, beam search, temperature sampling, top-$k$, and top-$p$ sampling. For a chosen strategy $\tau \in T$, we fix $x$ and $\theta$, generate $N$ responses $\{y^n\}_{n=1}^N$ under the selected sampling, and compute semantic entropy over the induced semantic clusters:
\begin{equation}
\begin{aligned}
U_{\text{dec}}(x, \theta, \tau)
&= -\sum_{c \in C} \hat{p}(c)\log \hat{p}(c), \\
\text{where}\quad
\hat{p}(c)
&= \frac{\sum_{y^n \in c} p(y^n \mid x, \theta, \tau)}
{\displaystyle\sum_{c' \in C}\sum_{y^n \in c'} p(y^n \mid x, \theta, \tau)}
\end{aligned}
\label{eq:decoding_uq}
\end{equation}
By comparing $U_{\text{dec}}$ across different $\tau \in T$, we can analyze how strongly the model's uncertainty depends on the choice of decoding policy.

Although these three sources of uncertainty are analyzed separately, they are not strictly orthogonal in practice. For example, an ambiguous prompt can also increase decoding variability by flattening the output distribution across multiple plausible interpretations. Thus, our framework should be viewed as a tool for diagnostic decomposition rather than an additive partition of a single total uncertainty score.

\section{Experiments}
In this section, we empirically validate our proposed uncertainty \textit{decomposition} framework. Our experiments are designed to answer two primary research questions:
% \begin{enumerate}[label=\textbf{RQ \arabic*:}]
%     \item Can our decomposed uncertainty scores effectively predict model failures across different tasks?
%     \item How do the dominant sources of uncertainty change with model scale and task type?
% \end{enumerate}

\medskip
\noindent\textbf{RQ 1:} Can our decomposed uncertainty scores effectively predict model failures across different tasks?

\medskip
\noindent\textbf{RQ 2:} How do the dominant sources of uncertainty change with model scale and task type?

\subsection{Experimental Setup}
In this section, we outline the components of our experimental design, including the datasets, models, and evaluation metrics used to validate our framework.

\subsubsection{Tasks and Datasets} 
We evaluate our framework on two distinct datasets to analyze uncertainty under different tasks, focusing on the LLM’s accuracy in zero-shot prediction. For factual question answering, we use TriviaQA, a dataset that requires models to provide factually correct responses, thereby testing their learned knowledge and ability to generate precise information. We also use GSM8K to evaluate the model on mathematical reasoning, as this dataset of grade-school math word problems assesses multi-step reasoning where errors may arise from either misinterpretation of the problem statement or flaws in the logical chain.

% We evaluate our framework on two distinct datasets to analyze uncertainty under different tasks. The experiments were conduct to evaluate LLM's accuracy in zero-shot prediction.

% \begin{itemize}
%     \item \textbf{Factual Question Answering:} We use TriviaQA, a challenging dataset that requires models to respond with answers that are factually correct. Success on this task depends on the model's learned knowledge. 
%     \item \textbf{Mathematical Reasoning:} For evaluating uncertainty in mathematical reasoning, we use GSM8K dataset which contains grade-school math word problems. This task tests multi-step reasoning, where errors can arise from misinterpretation of the problem or flaws in the logical chain.
% \end{itemize}

\subsubsection{Models} 
We conduct experiments across a range of model families and sizes, including Llama 3 (8B)~\cite{grattafiori2024llama} and Gemma 3 (270M, 1B, 4B, 12B and 27B)~\cite{team2025gemma}. Input and decoding based uncertainty estimation were tested on all these models, but model based uncertainty was only tested on Llama 3 8B and Gemma 3 27B.

\subsubsection{Implementation Details}

Below we describe the implementation details for each uncertainty component.

\textbf{Input Ambiguity (\(U_{\text{input}}\)).}
For each prompt, we generate \(K=5\) semantically similar paraphrases using GPT-5-nano. For each paraphrase, we obtain one response from the target LLM using greedy decoding. We then compute semantic entropy over the resulting set of responses using bidirectional entailment-based semantic clustering, as described in equation~\ref{eq:input_uq}.

\textbf{Knowledge Gaps (\(U_{\text{knowledge}}\)).}
We construct an ensemble of \(M=5\) model realizations for Llama 3 and Gemma 3 models by training LoRA adapters with different random seeds on the training split of the corresponding dataset. For a given prompt, we generate one response from each ensemble member using greedy decoding and compute semantic entropy over the ensemble outputs, as described in equation~\ref{eq:model_uq}.

\textbf{Decoding Randomness (\(U_{\text{dec}}\)).}
For a fixed decoding strategy \(\tau\), we generate \(N=5\) responses for each prompt using different random seeds and compute semantic entropy over these repeated generations, as described in equation~\ref{eq:decoding_uq}. We evaluate this separately for multiple practical decoding strategies. Specifically, greedy decoding uses a single deterministic decoding path; beam search uses 5 beams with length penalty 1.0 and early stopping; temperature sampling uses temperature \(0.7\); top-\(k\) sampling uses \(k=50\); and top-\(p\) sampling uses \(p=0.9\). Unless otherwise stated, the tabulated \(U_{\text{dec}}\) results are reported using temperature sampling.

\subsubsection{Evaluation Metrics}
To assess how effectively each decomposed uncertainty component predicts model failures (hallucinations), we cast the problem as a binary classification task that distinguishes correct from incorrect model outputs. The correctness of a generation is determined using a fuzzy matching approach, where an output is labeled correct if its Rouge-L score~\cite{lin2004automatic}, which measures the length of the longest common subsequence with respect to the reference answer, is greater than or equal to 0.3.

Once correctness is established, we evaluate predictive performance using the Area Under the Receiver Operating Characteristic curve (AUROC), where higher values indicate stronger alignment between uncertainty scores and actual model errors. In addition, we report the Expected Calibration Error (ECE) to quantify how well the uncertainty scores correspond to true error rates.

% We evaluate the effectiveness of each decomposed uncertainty component at predicting model failures (hallucinations). We frame this as a binary classification task where the goal is to distinguish between correct and incorrect model responses. The performance is measured using the Area Under the Receiver Operating Characteristic curve (AUROC). A higher AUROC indicates that the uncertainty score is more predictive of model failure. We also report the Expected Calibration Error (ECE) to assess how well the uncertainty scores reflect the true error rates.

\begin{table*}[t]
\centering
\caption{Failure prediction performance (AUROC) of each uncertainty component on TriviaQA (fact-retrieval) and GSM8K (reasoning). Higher values indicate a stronger ability to predict incorrect model responses. The results show that the most uncertainty source is task-dependent.}
\label{tab:uq_results}
\begin{tabular}{l l cc cc cc}
\toprule
\textbf{Dataset} & \textbf{Model} 
  & \multicolumn{2}{c}{\textbf{Input UQ}} 
  & \multicolumn{2}{c}{\textbf{Knowledge UQ}} 
  & \multicolumn{2}{c}{\textbf{Decoding UQ}} \\
\cmidrule(lr){3-4} \cmidrule(lr){5-6} \cmidrule(lr){7-8}
 & & AUROC & ECE & AUROC & ECE & AUROC & ECE \\
\midrule
\multirow{2}{*}{TriviaQA} 
  & Llama 3 (8B) & 0.705 & 0.340 & 0.499 & 0.513 & 0.731 & 0.364 \\
  & Gemma 3 (27B) & 0.761 & 0.223 & 0.498 & 0.514 & 0.636 & 0.458 \\
\midrule
\multirow{2}{*}{GSM8K} 
  & Llama 3 (8B) & 0.518 & 0.926 & 0.598 & 0.810 & 0.533 & 0.843 \\
  & Gemma 3 (27B) & 0.334 & 0.920 & 0.500 & 0.827 & 0.383 & 0.861 \\
\bottomrule
\end{tabular}
\end{table*}

\begin{figure*}[t]
\centering
\begin{subfigure}{0.28\linewidth}
    \centering
    \includegraphics[width=\linewidth]{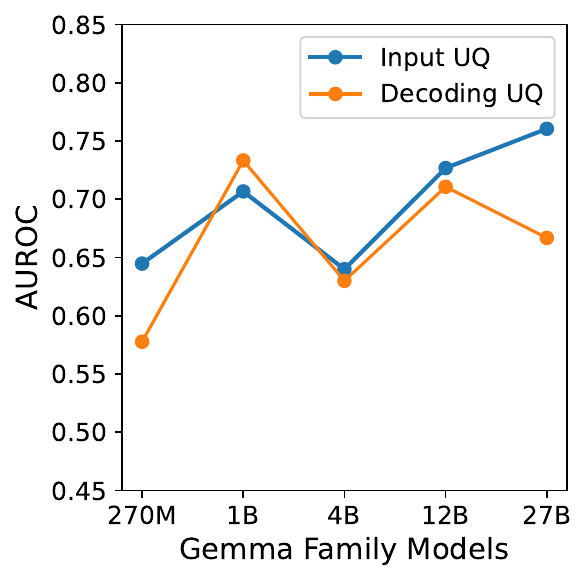}
    \caption{}
    \label{fig:uq_scaling}
\end{subfigure}
\hfill
\begin{subfigure}{0.68\linewidth}
    \centering
    \includegraphics[width=\linewidth]{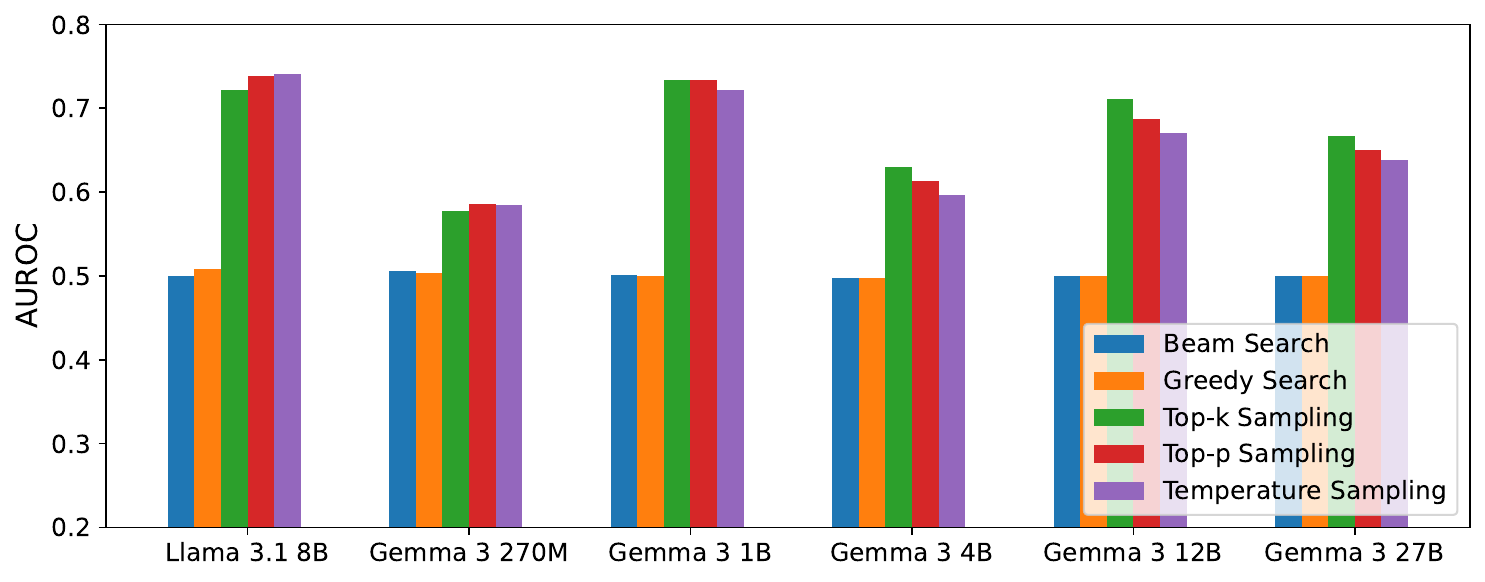}
    \caption{}
    \label{fig:uq_decoding_comparision}
\end{subfigure}
\caption{(a) Failure prediction performance (AUROC) of Input and Decoding uncertainty across the Gemma 3 model family on TriviaQA. As models scale, input ambiguity becomes a more reliable predictor of failure. (b) Comparison of failure prediction AUROC for Decoding Uncertainty when calculated using different decoding strategies. Stochastic methods (e.g., Top-k, Top-p) are significantly more effective at revealing uncertainty than deterministic ones (e.g., Greedy).}
\label{fig:uq_figures}
\end{figure*}

\begin{figure*}[t]
\centering
\begin{subfigure}{0.48\linewidth}
    \centering
    \includegraphics[width=\linewidth]{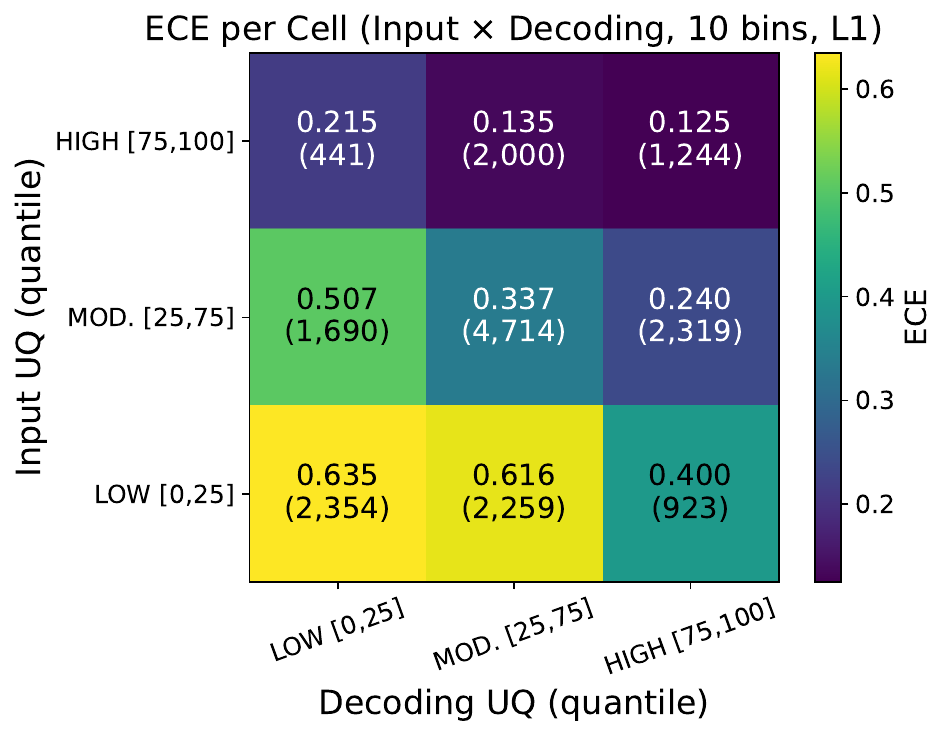}
    \caption{}
    \label{fig:uq_1}
\end{subfigure}
\hfill
\begin{subfigure}{0.48\linewidth}
    \centering
    \includegraphics[width=\linewidth]{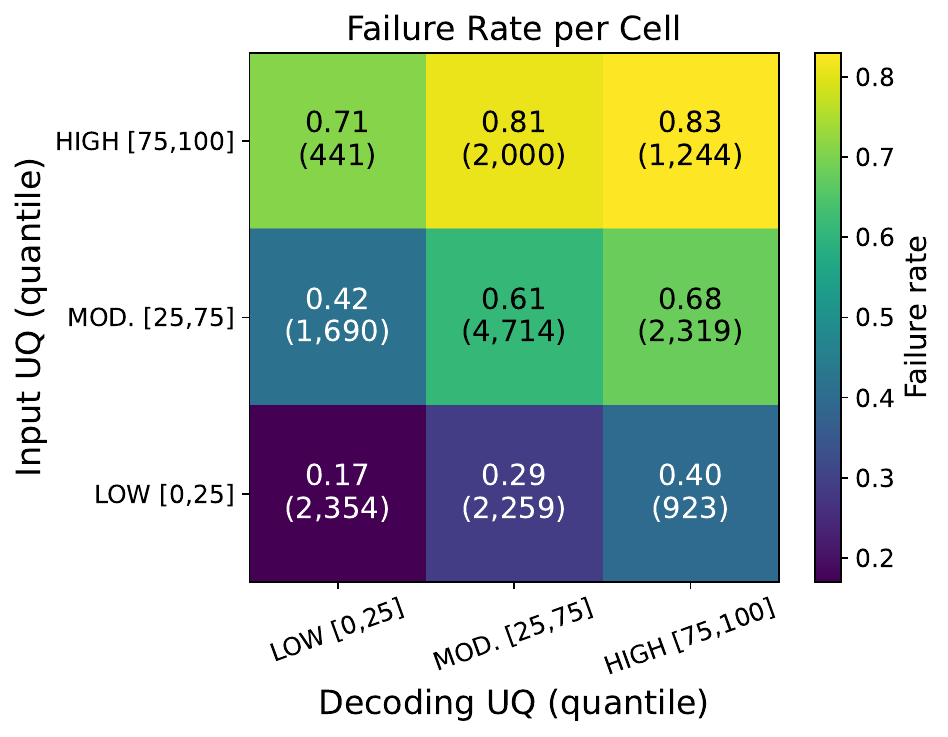}
    \caption{}
    \label{fig:uq_2}
\end{subfigure}
\caption{Joint analysis of Input Ambiguity ($U_{\text{input}}$) and Decoding Randomness ($U_{\text{dec}}$) on TriviaQA, partitioned by uncertainty quantiles. The heatmaps reveal an important insight about overconfidence: while the failure rate (b) increases with uncertainty, the model is most poorly calibrated (highest ECE in a) when it appears most confident (low uncertainty)}
\label{fig:uq_figures_2}
\vspace{-5pt}
\end{figure*}

\subsection{Disentangling Uncertainty for Failure Detection}
Table~\ref{tab:uq_results} presents the AUROC and ECE for each uncertainty component. The results demonstrate that the effectiveness of uncertainty decomposition is strongly \textbf{task-dependent}. On TriviaQA, both input ambiguity ($U_{\text{input}}$) and decoding randomness ($U_{\text{dec}}$) provide meaningful failure signals. For Llama~3~(8B), $U_{\text{dec}}$ is the strongest predictor (AUROC 0.731), suggesting that uncertainty about factual recall often manifests as variability under stochastic generation. In contrast, for Gemma~3~(27B), $U_{\text{input}}$ is most predictive (AUROC 0.761), indicating that failures in bigger models are driven more by sensitivity to how the question is phrased than by sampling noise.

On GSM8k, although all uncertainty components yield weak failure prediction signals, we observe that knowledge-based uncertainty is comparatively less degraded than input or decoding-based uncertainty. This suggests that reasoning failures are less driven by ambiguity or sampling variability, and more by confident but incorrect internal reasoning trajectories.

\subsection{Analysis of Scaling and Decoding Effects}
To answer RQ2, we analyze how uncertainty sources evolve with model scale and decoding choices. Figure~\ref{fig:uq_figures}(a) illustrates the performance of input and decoding uncertainty for the Gemma 3 model family on TriviaQA. We observe no clear monotonic trend; the predictive power of both uncertainty types fluctuates with model size. However, a notable pattern emerges: for smaller models (1B), decoding uncertainty is a stronger predictor, while for larger models (12B, 27B), input ambiguity becomes the more reliable signal. This reinforces our finding from Table~\ref{tab:uq_results}: as models grow, their sensitivity to input phrasing becomes a more prominent failure mode than simple generation variability.

Figure~\ref{fig:uq_figures}(b) explores the impact of different decoding strategies on uncertainty-based failure detection. A consistent and striking pattern emerges across all models: stochastic decoding methods (Top-k, Top-p, and Temperature Sampling) yield significantly higher AUROC scores than deterministic methods (Beam Search, Greedy Search). This demonstrates that allowing the model to explore a diverse set of potential answers is important for revealing its underlying uncertainty. Deterministic methods, which force the model to commit to a single path, can mask this uncertainty, often leading to confidently incorrect answers.

% As model size increases from 270M to 27B, $U_{\text{model}}$ consistently decreases, suggesting that larger models possess more comprehensive knowledge, leading to less internal disagreement. Conversely, $U_{\text{input}}$ remains relatively stable, indicating that sensitivity to prompt phrasing is a persistent challenge even for larger models.

% Figure~\ref{fig:uq_figures}(b) explores the impact of different decoding strategies on $U_{\text{dec}}$. As expected, greedy decoding produces minimal uncertainty, as it is deterministic. In contrast, increasing the temperature in sampling progressively raises the decoding uncertainty, creating a more diverse set of outputs. This result confirms that $U_{\text{dec}}$ directly reflects the stochasticity introduced at generation time, which can be tuned for different applications (e.g., high temperature for creative tasks, low for factual ones). Together, these analyses demonstrate that our framework provides a granular and interpretable view of model uncertainty.

\subsection{Interaction of Uncertainty Sources}
To better understand how different uncertainty sources interact, we performed a joint analysis of input ambiguity and decoding randomness. We partitioned the TriviaQA test set into a 3x3 grid based on low, moderate, and high quantiles of $U_{\text{input}}$ and $U_{\text{dec}}$. We then computed the average model failure rate and ECE within each cell.

Figure~\ref{fig:uq_figures_2}(b) shows a clear and intuitive trend: the model's failure rate increases monotonically with both input and decoding uncertainty. The lowest failure rate (0.17) occurs when both uncertainty scores are low, while the highest rate (0.83) occurs when both are high. This confirms that both components are meaningful indicators of correctness, and their combined effect is even stronger.

However, Figure~\ref{fig:uq_figures_2}(a) reveals a more surprising relationship with calibration. The model is most poorly calibrated (highest ECE of 0.635) when it appears most confident (low input and decoding uncertainty). Conversely, it is best calibrated (lowest ECE of 0.125) when it is most uncertain. This suggests that the model is often underconfident. When the model signals low uncertainty on both axes, it is only wrong 17\% of the time, but its confidence level is disproportionately low, leading to poor calibration. This highlights a critical failure mode: the model's confidence is least trustworthy precisely when it should appear most certain.

\section{Discussion: Actionable Uncertainty Decomposition}
\label{sec:discussion}

Uncertainty decomposition is valuable not only for measuring confidence but also for understanding why LLMs fail in the first place. When uncertainty is represented by a single scalar score, it only indicates that the model is unsure, without revealing the underlying reason. As a result, the only practical response is often abstention or fallback generation. In contrast, separating uncertainty into interpretable components helps diagnose the source of failure. For example, some failures arise because the prompt itself is ambiguous and admits multiple valid interpretations. Recent work shows that LLMs can improve reliability by detecting such underspecified queries and asking clarifying questions before generating a response~\cite{li2025questbench, yang2025prompts}. 

Other failures can originate from gaps in the model’s internal knowledge. Retrieval-based systems address this by augmenting the model with external information when parametric knowledge is insufficient. Self-Routing RAG~\cite{wu2025self}, for example, uses uncertainty signals to decide whether a query should be answered using the model’s internal knowledge or through external retrieval. However, recent analysis argues that standard predictive entropy fails to capture knowledge-related uncertainty in retrieval-augmented pipelines~\citep{soudani2025uncertainty}. This further supports the importance of identifying the source of uncertainty.

Uncertainty can also arise during the generation process itself. In structured tasks such as code generation, reliability can depend strongly on token choices during decoding. Frameworks such as AdaDec therefore monitor token-level entropy and trigger additional search or reranking when decoding uncertainty becomes high~\cite{he2025towards}. In summary, these works illustrate how decomposing uncertainty turns it into a practical signal for identifying model failures and improving LLM behavior, enabling targeted interventions such as clarification, retrieval, or adaptive decoding.

% On the other hand, Self-Routing RAG~\cite{wu2025self} uses uncertainty to show that when there is gaps in the model’s parametric knowledge, their selective retrieval methods dynamically decide whether external retrieval should be invoked or whether the model should rely on its internal knowledge. Similarly, analyses of uncertainty in retrieval-augmented systems highlight that traditional predictive entropy alone may fail to capture when parametric knowledge is insufficient, underscoring the need to isolate knowledge-driven uncertainty signals \citep{soudani2025uncertainty}.

% Additionally, recent studies highlight that in structured domains like code generation, output reliability is highly sensitive to token selection during the decoding phase. By monitoring decoding uncertainty, systems can dynamically adjust their generation strategies at runtime. For instance, the AdaDec framework leverages token-level entropy to trigger supplementary search or reranking mechanisms only when uncertainty peaks~\cite{he2025towards}. In summary, these research demonstrate how isolating uncertainty into actionable components provides a practical foundation for engineering adaptive, self-correcting LLM systems.

\section{Conclusion}
% Each component is formalized as a categorical distribution over semantic equivalence classes of responses, enabling direct comparison via a common semantic entropy measure. Through systematic evaluation across fact-retrieval and reasoning tasks, we show that the dominant source of uncertainty is \textbf{strongly task-dependent}, with input ambiguity being most predictive for factual queries, while standard uncertainty signals exhibit limited discriminative power for reasoning tasks. Our findings challenge the common practice of relying on a single uncertainty estimate and highlight fundamental differences in how semantic uncertainty manifests across task types.

We present \textbf{a unified framework for decomposing LLM uncertainty} into three distinct \textbf{semantic (not probabilistic)} components: input-induced, knowledge-induced, and decoding-induced uncertainty. Each component is formalized as a categorical distribution over semantic equivalence classes of responses, enabling direct comparison via a common semantic entropy measure. Through systematic evaluation across fact-retrieval and reasoning tasks, we show that the \textbf{dominant source of uncertainty is task- and model-dependent}. On TriviaQA, uncertainty decomposition yields meaningful failure signals with smaller models exhibit stronger decoding-driven uncertainty, while larger models are more sensitive to prompt phrasing (input ambiguity). 
% Through systematic evaluation across fact-retrieval and reasoning tasks, we show that the dominant source of uncertainty is \textbf{strongly task- and model-dependent}, with input ambiguity being most predictive for factual queries, while standard uncertainty signals exhibit limited discriminative power for reasoning tasks.
Our findings challenge the common practice of relying on a single uncertainty estimate and highlight fundamental differences in how semantic uncertainty manifests across task and model types.

% We argued that this source-aware ``anatomy of uncertainty'' provides a more diagnostically useful alternative to monolithic uncertainty scores. Our experiments demonstrated the practical value of this approach, showing that the dominant source of uncertainty shifts predictably with task and model scale, offering clear signals for targeted interventions like prompt refinement or retrieval augmentation.

% We also uncovered a critical challenge: a model's confidence is often least reliable when it appears most certain, highlighting the dangers of confidence. Furthermore, we identified that complex reasoning failures can evade detection by our current methods, pointing to an important frontier for future research. By moving beyond a single score and providing a more granular understanding of \textit{why} a model is uncertain, this work represents a step toward building more transparent, reliable, and ultimately more trustworthy language technologies.

% ==================
% Mandatory and not counted in page limit
% ==================
\section*{Limitations}

While our framework provides a practical approach for decomposing uncertainty in LLMs, several aspects remain open for improvement. In particular, our method relies on practical operational proxies to estimate the different sources of uncertainty. For example, input-based uncertainty is estimated using a finite set of paraphrases rather than considering the full space of possible prompt variations. Similarly, knowledge-based uncertainty is approximated using a small ensemble of LoRA-adapted model realizations instead of performing full posterior inference over model parameters. Furthermore, our estimates depend on the quality of semantic clustering, which may be affected by errors in the underlying entailment model.

\section*{Acknowledgments and Disclosure of Funding}

This work was performed under the auspices of the U.S. Department of Energy by the Lawrence Livermore National Laboratory under Contract No. DE-AC52-07NA27344, Lawrence Livermore National Security, LLC. This material is based upon work supported by the U.S. Department of Energy, Office of Science, Advanced Scientific Computing Research Program. LLNL-CONF-2015712.

\bibliography{custom}

\newpage
\appendix

\section{Appendix}
\label{sec:appendix}

\subsection{Compute Resources}
All local experiments were conducted on a single NVIDIA H100 GPU (80\,GB HBM) running Ubuntu 20.04 with CUDA 11.8. This setup was used for both LoRA ensemble training and local HuggingFace-based inference for uncertainty estimation on TriviaQA and GSM8K. Paraphrase generation for input ambiguity estimation was performed separately through the GPT-5-nano API.

\subsection{Prompt Templates}

\begin{figure}[h]
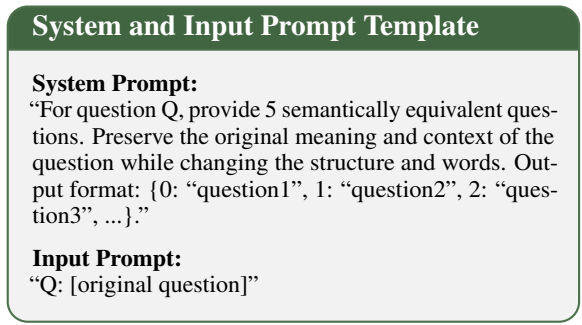

\centering
\begin{tcolorbox}[
    promptbox,
    title={System and Input Prompt Template},
    width=0.98\columnwidth
]
\small
\textbf{System Prompt:}

``For question Q, provide 5 semantically equivalent questions. Preserve the original meaning and context of the question while changing the structure and words. Output format: \{0: ``question1'', 1: ``question2'', 2: ``question3'', ...\}.''

\vspace{6pt}
\textbf{Input Prompt:}

``Q: [original question]''
\end{tcolorbox}
\caption{Prompt template used for paraphrase generation when estimating input ambiguity.}
\label{fig:paraphrase_prompt_box}
\end{figure}

For input ambiguity estimation, we generate semantically equivalent paraphrases of each input question using GPT-5-nano. Figure~\ref{fig:paraphrase_prompt_box} shows the instruction and input prompt template used for paraphrase generation.

\subsection{Decoding Hyperparameters}
For decoding-based uncertainty, we generate repeated outputs under a fixed decoding strategy and compute semantic entropy across these outputs. Unless otherwise stated, we use 5 generations per prompt. We evaluate multiple practical decoding strategies: greedy decoding, beam search with 5 beams, temperature sampling with temperature 0.7, top-$k$ sampling with $k=50$, and top-$p$ sampling with $p=0.9$. Beam search uses length penalty 1.0 and early stopping. The main results reported for $U_{\text{dec}}$ use temperature sampling unless specified otherwise.

\subsection{LoRA Ensemble Training Details}
To estimate knowledge-based uncertainty, we approximate multiple model realizations using an ensemble of LoRA-adapted models trained with different random seeds. For each base model, we train 5 LoRA adapters. The LoRA configuration uses rank $r=8$, $\alpha=32$, dropout 0.1, bias set to \texttt{none}, and target modules \texttt{q\_proj} and \texttt{v\_proj}. We train each adapter for 1 epoch with learning rate $2\times10^{-5}$, per-device batch size 4, and gradient accumulation steps 2. The maximum sequence length is set to 1024. For each input, one response is generated from each ensemble member using greedy decoding, and semantic entropy over these responses is used to estimate $U_{\text{knowledge}}$.

\begin{table}[t]
\centering
\caption{LoRA ensemble training hyperparameters.}
\label{tab:lora_hparams}
\begin{tabular}{lc}
\toprule
Hyperparameter & Value \\
\midrule
Number of LoRA models & 5 \\
LoRA rank $r$ & 8 \\
LoRA $\alpha$ & 32 \\
LoRA dropout & 0.1 \\
Bias & none \\
Target modules & q\_proj, v\_proj \\
Learning rate & $2\times10^{-5}$ \\
Batch size & 4 \\
Gradient accumulation & 2 \\
Epochs & 1 \\
Maximum sequence length & 1024 \\
\bottomrule
\end{tabular}
\end{table}

\subsection{Evaluation Metrics}
We evaluate each uncertainty component as a predictor of model failure. Let $a_i$ denote the reference answer for example $i$, let $\hat{a}_i$ denote the model's final answer, and let $u_i$ denote the corresponding uncertainty score (e.g., $U_{\text{input}}$, $U_{\text{knowledge}}$, or $U_{\text{dec}}$).

To determine whether a prediction is correct, we use bidirectional semantic equivalence based on a natural language inference (NLI) model. Specifically, an answer is marked correct only if the reference answer entails the generated answer and the generated answer also entails the reference answer. Formally, the correctness label is defined as,
\begin{equation}
\begin{aligned}
z_i = \mathbf{1}\!\Big[
&\, p_{\mathrm{NLI}}(a_i \Rightarrow \hat{a}_i) \ge \gamma \\
&\land\; p_{\mathrm{NLI}}(\hat{a}_i \Rightarrow a_i) \ge \gamma
\Big]
\end{aligned}
\end{equation}
where $p_{\mathrm{NLI}}(\cdot \Rightarrow \cdot)$ denotes the entailment probability returned by the NLI model and $\gamma=0.5$ in our experiments. We then define the failure label as,
\begin{equation}
f_i = 1 - z_i,
\end{equation}
so that $f_i=1$ indicates an incorrect response and $f_i=0$ indicates a correct one.

\paragraph{AUROC.}
To evaluate how well an uncertainty score separates failures from correct responses, we compute the Area Under the Receiver Operating Characteristic curve (AUROC). AUROC measures the probability that a randomly chosen failed example receives a higher uncertainty score than a randomly chosen correct example. It can be written as,
\begin{equation}
\mathrm{AUROC}
=
\frac{1}{|\mathcal{P}||\mathcal{N}|}
\sum_{i \in \mathcal{P}}
\sum_{j \in \mathcal{N}}
\mathbf{1}[u_i > u_j],
\end{equation}
where $\mathcal{P}=\{i : f_i=1\}$ is the set of failed examples and $\mathcal{N}=\{j : f_j=0\}$ is the set of correct examples. Higher AUROC indicates that the uncertainty score is more effective at ranking incorrect generations above correct ones.

\paragraph{Expected Calibration Error (ECE).}
In addition to ranking performance, we evaluate calibration to measure whether larger uncertainty values correspond to higher empirical failure rates. Since ECE requires a confidence-like quantity in $[0,1]$, we first map each uncertainty score $u_i$ to a failure-confidence score $\hat{p}_i \in [0,1]$ using a monotonic normalization. We then partition predictions into $B$ bins according to $\hat{p}_i$. For bin $b$, let $\mathcal{B}_b$ denote the set of assigned examples. The average predicted failure confidence and empirical failure rate in that bin are,
\begin{equation}
\begin{aligned}
\operatorname{conf}_b
&=
\frac{1}{|\mathcal{B}_b|}
\sum_{i \in \mathcal{B}_b} \hat{p}_i, \\
\operatorname{err}_b
&=
\frac{1}{|\mathcal{B}_b|}
\sum_{i \in \mathcal{B}_b} f_i.
\end{aligned}
\end{equation}
The Expected Calibration Error is then defined as,
\begin{equation}
\mathrm{ECE}
=
\sum_{b=1}^{B}
\frac{|\mathcal{B}_b|}{n}
\left|
\mathrm{err}(\mathcal{B}_b)
-
\mathrm{conf}(\mathcal{B}_b)
\right|,
\end{equation}
where $n$ is the total number of examples. Lower ECE indicates better calibration, meaning that the uncertainty-derived confidence values more closely match the observed failure frequencies.

\end{document}